\documentclass[sigconf]{acmart}
\usepackage{etoolbox}
\makeatletter
\patchcmd{\NAT@test}{\else \NAT@nm }{\else \NAT@nmfmt{\NAT@nm}}{}{}
\makeatother

\copyrightyear{2024}
\acmYear{2024}
\setcopyright{rightsretained}
\acmConference[FAccT '24]{The 2024 ACM Conference on Fairness, Accountability, and Transparency}{June 3--6, 2024}{Rio de Janeiro, Brazil}
\acmBooktitle{The 2024 ACM Conference on Fairness, Accountability, and Transparency (FAccT '24), June 3--6, 2024, Rio de Janeiro, Brazil}\acmDOI{10.1145/3630106.3659010}
\acmISBN{979-8-4007-0450-5/24/06}
\usepackage{tikz}
\usetikzlibrary{shapes,decorations,arrows,calc,arrows.meta,fit,positioning}
\tikzset{
    -Latex,auto,node distance =1 cm and 1 cm,semithick,
    point/.style = {circle, draw, inner sep=0.04cm,fill,node contents={}},
    bidirected/.style={Latex-Latex,dashed},
    latent/.style={dashed},
    inextricable/.style={draw,double},
}

\usepackage{graphicx,subcaption}
\usepackage{breakurl}

\begin{document}

\title{Unlawful Proxy Discrimination: A Framework for Challenging Inherently Discriminatory Algorithms}

\author{Hilde Weerts}
\affiliation{%
  \institution{Eindhoven University of Technology}
  \country{The Netherlands}}
\email{h.j.p.weerts@tue.nl}

\author{Aislinn Kelly-Lyth}
\affiliation{%
  \institution{Blackstone Chambers}
  \country{United Kingdom}}
\email{aislinnkelly-lyth@blackstonechambers.com}

\author{Reuben Binns}
\affiliation{%
  \institution{University of Oxford}
  \country{United Kingdom}}
\email{reuben.binns@cs.ox.ac.uk}

\author{Jeremias Adams-Prassl}
\affiliation{%
  \institution{University of Oxford}
  \country{United Kingdom}}
\email{jeremias.adams-prassl@law.ox.ac.uk}

\renewcommand{\shortauthors}{}

\begin{abstract}
Emerging scholarship suggests that the EU legal concept of direct discrimination - where a person is given different treatment on grounds of a protected characteristic - may apply to various algorithmic decision-making contexts. This has important implications: unlike indirect discrimination, there is generally no `objective justification' stage in the direct discrimination framework, which means that the deployment of directly discriminatory algorithms will usually be unlawful per se. In this paper, we focus on the most likely candidate for direct discrimination in the algorithmic context, termed \emph{inherent} direct discrimination, where a proxy is inextricably linked to a protected characteristic.
We draw on computer science literature to suggest that, in the algorithmic context, `treatment on the grounds of' needs to be understood in terms of two steps: proxy \emph{capacity} and proxy \emph{use}. Only where both elements can be made out can direct discrimination be said to be `on grounds of' a protected characteristic. We analyse the legal conditions of our proposed proxy capacity and proxy use tests. Based on this analysis, we discuss technical approaches and metrics that could be developed or applied to identify inherent direct discrimination in algorithmic decision-making.
\end{abstract}





\maketitle

\section{Introduction}
The challenges posed by bias in algorithmic systems have received extensive scrutiny in computer science and legal scholarship. Discrimination is unlawful in both United States (US) and European Union (EU) law when an individual is either treated less favourably because of a protected characteristic (known as \textit{disparate treatment} or \textit{direct discrimination}, respectively) or when the application of a seemingly neutral provision, criterion, or practice puts certain individuals at a particular, unjustifiable, disadvantage (\textit{disparate impact} or \textit{indirect discrimination}) \citep{fredman2022discrimination}.

A rich literature working to connect legal non-discrimination frameworks to statistical fairness metrics~\citep{barocas2016big,hacker2018teaching,xenidis2019eu,wachter2021fairness,weerts2023algorithmic} has primarily focused on disparate impact or indirect discrimination as the key avenue to challenge automated bias~\citep{binns2023legal}. This approach has potential advantages, particularly in the context of complex systems: focusing on their effects avoids difficult technical questions, from causation to interpretability. But there are also clear drawbacks: disparate impact and indirect discrimination will not be unlawful where the deployer of a system can provide a proportionate justification. A ride-hailing company might argue, for example, that even though a facial recognition system has higher error rates for certain populations, its use is the only feasible way of ensuring passenger safety.

This focus is (doctrinally) justified in US law: a clear requirement of intentional discrimination or explicit protected characteristic-based classification severely limits the scope of disparate treatment. In consequence, most scholars have assumed that unless variables have been purposefully selected to disadvantage protected groups (which is likely to be rare in practice), excluding protected characteristics from the features of a machine learning model is necessary and sufficient for avoiding disparate treatment. In EU law, on the other hand, the picture is more complex. As an emerging body of work has argued, the scope of direct discrimination is significantly broader \citep{adamsprassl2022directly,binns2023legal}. In the absence of any requirements for intent or moral culpability, at least certain types of algorithmic bias are caught in the scope of direct discrimination – with significant practical implications: treating an individual less favourably \textit{`on grounds of'} of a protected characteristic is near-universally unlawful; only very limited justifications can be adduced. A reframing of algorithmic discrimination as direct thus has clear legal attractions. However, the ensuing shift from effects-based to reason-focused treatment raises significant technical questions. How can a claimant show the required connection between their protected characteristic and the ensuing adverse treatment meted out by a complex algorithmic system? 

Previous work on algorithmic direct discrimination has identified two strands in the relevant EU case law \citep{adamsprassl2022directly}.\footnote{We have adopted the approach, also taken by \citet{adamsprassl2022directly}, of leveraging the detailed reasoning of the UK courts to understand the EU jurisprudence. We have done so in light of the close relationship between the two regimes, which have developed to mirror one another over a period of decades.}
 First, a decision may be \textit{inherently discriminatory} because it was made using a criterion which is inextricably linked to a protected characteristic. In \textit{Dekker}~\citep{dekker}, for example, the Court regarded pregnancy as inextricably linked to sex and found a decision based on pregnancy to be inherently discriminatory on grounds of sex. This form of direct discrimination closely resembles notions of \textit{proxy discrimination} in the algorithmic fairness literature \citep[e.g.,][]{barocas2016big,prince2019proxy,tschantz2022proxy}, which occurs when one or more input features of a protected characteristic act as a stand-in for a protected characteristic. Second, a decision may be made through \textit{subjectively discriminatory} mental processes. This will occur when a protected characteristic plays a role (whether consciously or subconsciously) in a decision maker’s less favourable treatment of a person \citep{CHEZ,Firma}. Given the emphasis on mental processes in human decision-making, subjective discrimination raises complex questions when translated into the algorithmic context. For example, case law on subconscious subjective discrimination has been developed intuitively by judges who generally ask on the facts whether a given (human) decision was `on grounds of’ a protected characteristic. As a result, the jurisprudence does not establish any clear test for subjective discrimination of a type which could be translated into a machine context.\footnote{More generally, identifying a framework for the assessment of implicit algorithmic bias will be challenging in circumstances where machine learning models are trained on data from an inherently unequal world.} In this work, we therefore focus on the first subcategory: inherent direct discrimination.

The primary contribution of this work is a framework to guide both litigants and developers in identifying unlawful inherent discrimination. Taking inspiration from notions of proxy discrimination proposed in the computer science literature \citep{tschantz2022proxy}, we suggest that the measurement of inherent discrimination can be subdivided into two distinct problems: \textit{proxy capacity} and \textit{proxy use}. We suggest that to produce an inherently discriminatory outcome, a model must (i) contain a criterion (or criteria) which is (or are) inextricably linked to a protected characteristic (proxy \textit{capacity}) and (ii) apply that criterion (or criteria) with the result that an individual is treated less favourably (proxy \textit{use}).

Before turning to a detailed development of this framework, an important preliminary point should be raised. Translating complex legal requirements into strict mathematical tests is an exercise fraught with risk, not least given the danger of legal nuance being erased in the process \citep{watkins2022fourfifths}. That risk is particularly pronounced in the present context, given that the Court of Justice of the European Union (CJEU) has approached the concept of inherently discriminatory criteria in an intuitive, and consequently sometimes inconsistent, manner.

Our proposed framework should therefore not be understood as an exhaustive technical definition of directly discriminatory algorithmic systems: some forms of proxy discrimination will not be unlawful in this way; conversely a model may still be directly discriminatory where the proxy capacity and/or proxy use framework is not satisfied. This is driven both by inconsistencies in the judicial interpretation of inherent discrimination, and the existence of additional categories, notably subjective direct discrimination, as well as other potential forms of direct discrimination as yet to be defined by the courts (some of which may be novel to the algorithmic context \citep{binns2023legal}). These are beyond the scope of the present paper. Our purpose in this work is to map out the \emph{clearest} basis on which a claimant might argue successfully that an algorithm is inherently directly discriminatory within the current legal framework.

Our focus on a ‘core’ case of unlawful inherent discrimination consequently means that our framework is not designed to be translated into a test which, if met, would indicate that a model is free from any risk of inherent direct discrimination. Rather, it is intended to be of use to prospective claimants and to algorithmic developers seeking to identify preliminary ‘red flags’ in their models. The development of a framework is important, given the significant potential for undetected inherent algorithmic discrimination. This is particularly true in the context of machine learning, where machine learning practitioners may not know that certain features are linked to protected characteristics, especially where such a link only arises when multiple features are combined.

In developing this analysis, the remainder of this paper is structured as follows. Section~\ref{sec:algorithmicproxydiscrimination} begins with a discussion of existing work on algorithmic proxy discrimination. In Section~\ref{sec:legalanalysis}, we then turn to the legal background to our proposed \textit{proxy capacity} and \textit{proxy use} tests. Based on this analysis, Section~\ref{sec:technicalanalysis} covers technical approaches and metrics that could be developed or applied. The technical and legal implications of this approach are discussed in Section~\ref{sec:discussion}. Section~\ref{sec:conclusion} concludes.
 
\section{Algorithmic Proxy Discrimination}
\label{sec:algorithmicproxydiscrimination}
Before articulating why we believe inherent direct discrimination can be understood as a type of proxy discrimination, it is necessary first to summarise some key definitions and distinctions around this concept, as established in the existing algorithmic fairness literature on which our argument builds. \textit{Proxy discrimination} occurs when a facially neutral feature is used in a predictive model as a stand-in for a protected characteristic \citep{barocas2016big,prince2019proxy,tschantz2022proxy}. Historically, the term proxy discrimination was used in the US to refer to forms of intentional discrimination in human decision-making, where the use of a proxy was motivated by its association with protected group membership \citep{prince2019proxy}. Perhaps the most well-known example is redlining, a practice in which services were denied to residents of particular neighbourhoods considered "risky" on the basis of racial or ethnic composition. With the proliferation of algorithmic decision-making, concerns regarding unintentional forms of proxy discrimination have become more prominent. It is widely acknowledged that the exclusion of protected characteristics from predictive models, where predictions produce particular benefits or burdens, is insufficient to avoid unfavourable treatment \citep{dwork2012fairness}. Specifically, if a target variable is associated with a protected characteristic, any sufficiently accurate predictive model will reproduce this association. Even if a protected characteristic is excluded from the training data, it can often still be inferred (implicitly) through correlations with other features in the training data. For example, in cities where neighbourhoods are segregated along ethnic lines, postal code can be a proxy for ethnicity. While the term `proxy discrimination' frequently occurs in the literature, (implicit) definitions vary. To clarify the different concerns captured by proxy discrimination, we leverage the framework proposed by \citet{tschantz2022proxy}, which suggests that proxy discrimination consists of two primary elements: proxy capacity and proxy use.

\subsection{Proxy Capacity}
\begin{figure}[ht]
    \centering
    \footnotesize
    \begin{subfigure}[t]{\linewidth}
    \centering
    \begin{tikzpicture}[node distance =0.85 cm and 0.85 cm]
        \node (a) [point, 
            label={[align=center]below:{$A$}}];
        \node (p) [point, right = of a, xshift=1cm,
            label={[align=center]below:{$P$}}];
        \node (u) [point, above = of a,
            label={[align=center]above:{$U$}}];  
        \path (u) edge (a);
        \path (u) edge (p);
    \end{tikzpicture}
    \caption{$A$ and $P$ are associated through confounder $U$ (via the path $A \leftarrow U \rightarrow P$), but $P$ is not a causal descendent of $A$.}
    \label{fig:kilbertusno}
    \end{subfigure}
    \qquad
    \begin{subfigure}[t]{\linewidth}
    \centering
    \begin{tikzpicture}[node distance =0.85 cm and 0.85 cm]
        \node (a) [point, 
            label={[align=center]below:{$A$}}];
        \node (p) [point, right = of a, xshift=1cm,
            label={[align=center]below:{$P$}}];
        \node (u) [point, above = of a,
            label={[align=center]above:{$U$}}];  
        \path (u) edge (a);
        \path (u) edge (p);
        \path (a) edge (p);
    \end{tikzpicture}
    \caption{$P$ is a causal descendent of $A$, indicating causal proxy capacity proposed by \citet{kilbertus2017avoiding}.}
    \label{fig:kilbertusyes}
    \end{subfigure}
    \caption{Two causal graphs that can produce statistical proxy capacity. }
    \label{fig:causalexplanation}
\end{figure}
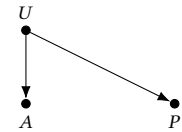
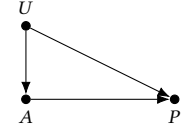

\textit{Proxy capacity} refers to the extent to which $A$, a variable that represents a protected characteristic, can be recreated from a proxy variable $P$. Both statistical and causal notions of proxy capacity have been proposed. Statistical measures of proxy capacity typically quantify some form of statistical association between $P$ and $A$, such as correlation. Causal notions of proxy capacity take into consideration the causal relationships between $P$ and $A$. Assumed causal relationships are typically modelled in the form of a directed acyclic graph,\footnote{In mathematics, a graph is a structure which denotes a set of objects ("nodes") and the relationships between them ("edges"). In a directed acyclic graph, relationships are directed explicitly from one node to another ("directed") and none of the edges form a directed cycle ("acyclic").} in which each node represents a variable and each directed edge the existence of a causal relationship between two variables \citep{pearl2016causal}. \citet{kilbertus2017avoiding} constrain the notion of 'proxy' to variables that are descendants of the protected characteristic in such a causal graph. To make clear the distinction between statistical and causal capacity we can consider the causal graphs depicted in Figure~\ref{fig:causalexplanation}. Assuming the causal structure in Figure~\ref{fig:kilbertusno}, both $A$ and $P$ are caused by $U$, but are otherwise not causally related. The causal graph in Figure~\ref{fig:kilbertusyes} depicts the same assumptions, in addition to the assumption that $A$ causes $P$. Both graphs depict a scenario in which proxy $P$ has statistical capacity for $A$, but only Figure~\ref{fig:kilbertusyes} adheres to the causal definition of proxy capacity put forward by \citet{kilbertus2017avoiding}.

\subsection{Proxy Use}
Some form of proxy capacity seems necessary to refer to a variable as a proxy for a protected characteristic. However, proxy capacity alone is insufficient to speak of proxy discrimination. For example, consider a face recognition data set. Modern machine learning algorithms would likely be able accurately to predict skin colour, which falls under the protection of race, from an image. While this can be problematic in many scenarios,\footnote{For example, \citet{spanton2022measuring} draw parallels with the pseudoscientific practice of physiognomy.} it does not imply that \textit{all} applications of facial recognition based on that data set actually use the capacity to proxy race in a way that \emph{disadvantages} members of a protected group (in this case, racial groups). This is captured by the second primary element of proxy discrimination, \textit{proxy use}, which considers the extent to which a proxy $P$ induces predictions $\hat{Y}$. For intrinsically interpretable models, proxy use can be deduced from the internal model structure. For example, in a linear regression model, the strength of the coefficient that corresponds to the proxy variable $P$ indicates proxy use. For more complex model classes, proxy use can be deduced via an interventional analysis that reveals the influence of a proxy $P$ on predictions $\hat{Y}$. Again, \citet{kilbertus2017avoiding} take a broader perspective on proxy use and define proxy discrimination to occur if a causal intervention on a proxy variable changes the predicted outcome. This notion of proxy discrimination considers not only the effect of changing the proxy variable in isolation but also the downstream effects that intervening on the proxy would have on the other input features, due to causal relationships between the proxy and those features.\footnote{\citet{kilbertus2017avoiding} follow the causal calculus ('do-calculus') paradigm as proposed by \citet{pearl2016causal}. Causal intervention in this example means that we assume a world in which we have changed education major to be a certain value, e.g., all applicants majored in computer science. Following causal calculus theory, such interventions can be purely hypothetical: it is not necessary to be able to execute the intervention in practice to make causal inferences. For example, imagine a linear regression model used for selecting resumes for a software engineering position that includes an applicant's education major and years of programming experience as input features. We are interested in potential gender proxy discrimination based on an applicant's major. As it turns out, the model has a high, positive coefficient for years of experience, but a negligible coefficient for major. Assuming that a person's major has a causal influence on years of experience (because students majoring in software engineering are programming during their studies, compared to others who are more likely to only begin after graduation), intervening on major would affect years of programming experience. The causal perspective on proxy use taken by \citet{kilbertus2017avoiding} could therefore consider this model to proxy discriminate on major -- even though the variable that represents major hardly affects an applicant's prospects.}

Regardless of exactly how proxies and their causal effects on $\hat{Y}$ are dealt with, the distinction between \emph{capacity} and \emph{use} is important. Many sets of features used in a model may have the capacity to act as a proxy for protected characteristics, without the model using that capacity when generating its outputs. As we will argue below, this key distinction separates (inherent) direct discrimination from indirect discrimination.

\subsection{What Makes A Proxy A Proxy?}
In addition to the two primary elements of proxy use and proxy capacity, some authors have further qualified algorithmic proxy discrimination based on the \textit{reason} a proxy acts as a proxy. 

In some cases, this qualification is implicit. For example, \citet{kilbertus2017avoiding} define a proxy as "a descendant of [the protected characteristic] in the causal graph we choose to label as a proxy", which could suggest a distinction between causal descendants that would be wrong to act upon (i.e., proxies) and others (i.e., non-proxies). Other authors are more explicit. Taking a legal perspective, \citet{prince2019proxy} view proxy discrimination as a subset of disparate impact (or, analogously, indirect discrimination) in which the proxy feature derives its predictive power from its association with a protected characteristic, which \citet{tschantz2022proxy} refers to as capacity-induced proxy discrimination. 

\citet{prince2019proxy} argue that to distinguish between proxy discrimination and other forms of indirect discrimination, we must consider the causal mechanisms that are responsible for proxy capacity (i.e., a relationship between $A$ and $P$) and proxy use (i.e., the extent to which predictions $\hat{Y}$ are induced by $P$). As machine learning models are designed to accurately predict target variable $Y$, the latter is highly related to the association between proxy $P$ and target variable $Y$. According to the definition put forward by \citet{prince2019proxy}, proxy discrimination occurs when the association between proxy $P$ and target outcome $Y$ is the result of a causal link between the protected characteristic $A$ and target outcome $Y$, possibly mediated by another unavailable, unmeasurable, or excluded variable. 

\begin{figure}[ht]
    \centering
    \footnotesize
    \begin{subfigure}[t]{\linewidth}
    \centering
    \begin{tikzpicture}[node distance =0.85 cm and 0.85 cm]
        \node (a) [point, 
            label={[align=center]below:{$A$\\HTT mutation}}];
        \node (p) [point, right = of a, xshift=1cm,
            label={[align=center]below:{$P$\\patient support group}}];
        \node (y) [point, above = of p,
            label={[align=center]above:{health outcomes\\$Y$}}];
        \path (a) edge (p);
        \path (a) edge (y);
    \end{tikzpicture}
    \caption{Example of proxy discrimination as defined by \citep{prince2019proxy}. $P$ is associated with $Y$, due to a direct causal link between $A$ and $Y$ (via path $Y \leftarrow A \rightarrow P$).}
    \label{fig:capacityinducedproxy}
    \end{subfigure}
    \qquad
    \begin{subfigure}[t]{\linewidth}
    \centering
    \begin{tikzpicture}[node distance =0.85 cm and 0.85 cm]
        \node (a) [point,
            label={[align=center]below:{$A$\\gender}}];
        \node (p) [point, right = of a, xshift=1cm,
            label={[align=center]below:{$P$\\{part-time work}}}];
        \node (y) [point, above = of p,
            label={[align=center]above:{crash\\$Y$}}];  
        \node (u) [point, above = of a,
            label={[align=center]above:{driver care\\$U$}}];  
        \path (a) edge (p);
        \path (u) edge (y);
        \path (a) edge (u);
    \end{tikzpicture}
    \caption{Example of proxy discrimination as defined by \citep{prince2019proxy}. $P$ is associated with $Y$, due to an indirect causal link between $A$ and $Y$ (via path $Y \leftarrow U \leftarrow A \rightarrow P$).}
    \label{fig:indirectcapacityinducedproxy}
    \end{subfigure}
    \qquad
    \begin{subfigure}[t]{\linewidth}
    \centering
    \begin{tikzpicture}[node distance =0.85 cm and 0.85 cm]
        \node (a) [point, 
            label={[align=center]below:{$A$\\{religion}}}];
        \node (p) [point, right = of a, xshift=1cm,
            label={[align=center]below:{$P$\\{vocabulary}}}];
        \node (y) [point, above = of p,
            label={[align=center]above:{qualifications\\$Y$}}];
        \node (u) [point, above = of a,
            label={[align=center]above:{level of education\\$U$}}];  
        \path (u) edge (a); 
        \path (u) edge (y);
        \path (u) edge (p);
    \end{tikzpicture}
    \caption{Example of indirect non-proxy discrimination, according to \citep{prince2019proxy}. $P$ is associated with $Y$, but the predictive power is not derived from its relationship with $A$ but its relationship with $U$ (via path $P \leftarrow U \rightarrow Y$).}
    \label{fig:indirect}
    \end{subfigure}
    \caption{Examples of causal graphs in which $A$ and $P$ are associated with $Y$.}
    \label{fig:pdd}
\end{figure}
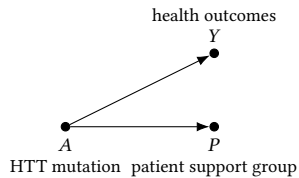
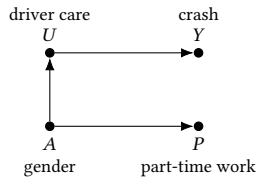
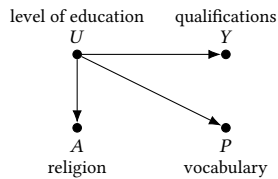

For example, Huntington's disease is directly caused by a mutation in the HTT gene, which affects healthcare outcomes. An insurer could hypothetically use membership of a patient support group as a proxy for the HTT mutation to predict health outcomes (Figure~\ref{fig:capacityinducedproxy}). Similarly, women tend to drive more safely than men, resulting in fewer car insurance claims (Figure~\ref{fig:indirectcapacityinducedproxy}). If more direct measures of driver care are unavailable, a model may use a facially neutral characteristic that proxies for gender, such as part-time work, to predict insurance claims. In contrast, \citet{prince2019proxy} argue that if a model uses a proxy feature that `fortuitously happens' to be correlated with a protected characteristic, this should not be viewed as proxy discrimination, but as a more general form of indirect discrimination. For example, we may assume that (1) religious affiliation is influenced by the level of education, where a higher level of education can cause a loss of religious belief \citep{zhang2018fairness,barocas2023fairness}, and (2) one's level of education affects one's vocabulary (Figure~\ref{fig:indirect}). A hiring algorithm could pick up on the vocabulary used in an application letter to predict qualifications. Due to the assumed relationship between level of education and religion, the use of vocabulary in a predictive model could disproportionately affect individuals with religious beliefs. According to the definition of \citet{prince2019proxy}, this would not be a case of proxy discrimination.

As discussed below, these qualifications and distinctions help clarify how the legal concept of inherent discrimination can be considered through the framework of proxy discrimination, without thereby collapsing the distinction between direct and indirect discrimination.

\section{Legal Conditions for Inherent Direct Discrimination} 
\label{sec:legalanalysis}
Having covered how the algorithmic fairness literature has qualified the nature of proxy discrimination, we can now explain why inherent discrimination in an algorithmic context can be viewed as a special case of algorithmic proxy discrimination, in which the proxy is not only associated with a protected characteristic but `inextricably linked' to it. As such, inherent discrimination as a form of direct discrimination in EU law shares several commonalities with notions of proxy discrimination in the algorithmic fairness literature. At the same time, there are important differences. Notions of proxy discrimination have been primarily motivated by, or analysed under, the disparate impact or indirect discrimination doctrines, resulting in an emphasis on discriminatory effects and potential justifications. In contrast, as a form of direct discrimination, legal conditions for inherent discrimination are reason-focused and generally do not leave room for justifications. Consequently, the emphasis shifts from potential justifications of discriminatory effects to establishing an `indissociable' or `inextricable' connection between a protected characteristic and a criterion.

Following the general characterisation of algorithmic proxy discrimination put forward in the previous section, we suggest that a successful case of inherent direct discrimination requires showing (1) \textit{proxy capacity}: there exists an `inextricable link' between a protected characteristic $A$ and its proxy $P$, and (2) \textit{proxy use}: a causal relationship between the proxy $P$ and the predicted outcome $\hat{Y}$ such that members of a protected group are treated less favourably.\footnote{Note that the causal relationship between proxy $P$ and predictions $\hat{Y}$ refers only to the predictive model - the causal relationship may or may not exist between $P$ and real-world outcome $Y$.}

The case law is not always explicit on why a criterion is `inextricably linked' to a protected characteristic, or how significant a criterion has to be in order for its application to constitute or result in less favourable treatment. In the remainder of this section, we explain how the courts have approached each of the two steps.

\subsection{Proxy Capacity}
\label{sec:legalcapacity}
What does it mean to speak of an `inextricable link' between a protected characteristic and its proxy? Many inextricably linked criteria recognised by the courts have some intuitive appeal. Pregnancy and sex, as mentioned above, is one example. While not a necessary or sufficient condition, the capacity for pregnancy is clearly linked to being assigned female at birth. Another example is marriage and sexual orientation in countries which prohibit same-sex marriage. If a couple is \textit{unable} to marry, a credit scoring model which uses marriage to predict creditworthiness will be inherently discriminatory on grounds of sexual orientation, as the CJEU recognised in \textit{Tadao} \citep{Tadao}.

What is less clear is where the boundary lies. How strong a relationship must there be between the protected characteristic $A$ and its proxy $P$ to speak of an `inextricable link’? In positing that some forms of proxy discrimination are best viewed as inherent direct discrimination, we are not claiming that all cases of proxy discrimination are inherent direct discrimination. The notion of inextricability thus needs to be able to pick out that subset of proxy discrimination which is (inherent) direct discrimination, without also including cases of indirect discrimination. 

In the United Kingdom (UK), judges have approached this challenge by holding that a criterion will only be inextricably linked to a protected characteristic if it ‘exactly corresponds’ with it \citep{Essop}. UK courts state that while it is not necessary for every person with the protected characteristic to be disadvantaged by the criterion used \citep{Coll}, it \textit{is} necessary that every person disadvantaged by the criterion have the protected characteristic. Put differently, the only people disadvantaged by the criterion must be those who fall within the defined protected group. 

The classic case that UK judges cite to make this point \citep{Essop, Ashers, Patmalniece} is \textit{James v Eastleigh }\citep{james}, a case in which free entry was provided to a swimming pool for those who had reached statutory retirement age. At the time, the statutory retirement age was 60 for women, but 65 for men. As such, women aged 60 to 64 could enter the pool for free, but men aged 60 to 64 could not: there was an ‘exact correspondence’ between those disadvantaged by the ‘free for retirees’ policy and those with the protected characteristic of being male. It did not matter that the policy only disfavoured a subgroup of men (i.e. those between the ages of 60 and 64), as long as \textit{only} \textit{men} were disadvantaged by the policy. 

This test works well when a criterion is based on an expressly discriminatory policy (even when that policy lies outside the organisation's own remit, like the national statutory retirement age policy), because such situations permit bright lines to be drawn. Where the relationship between the criterion and the protected characteristic arises instead from `organic' social conditions, it is less apt. In the recent case of \textit{WABE }\citep{wabe}, for example, the CJEU considered whether a rule which prohibits the wearing of all visible political, philosophical or religious signs in the workplace would constitute inherent direct discrimination. The CJEU answered this question in the negative, since "every person may have a religion or belief". The Court went on to find, however, that a prohibition which is limited to the wearing of \textit{conspicuous, large-sized}  signs of political, philosophical or religious beliefs \textit{is} liable to constitute direct discrimination, because certain religions "require the wearing of a large-sized sign, such as a head covering".

The approach developed by the CJEU is thus broader than the UK courts’ ‘exact correspondence’ test: not \textit{all} individuals who wear conspicuous, large-sized signs of political, philosophical or religious beliefs do so for religious reasons: indeed, they might be choosing conspicuously to display a \textit{political} or \textit{philosophical} view (\textit{res ipsa loquitur}). In other words, the criterion was liable to disadvantage (some) individuals without the protected characteristic; but the Court was alive to the particular proximity between the criterion and the requirements of certain religions.

In short, the courts have not expressly set out a reliable test with clear boundaries. How, then, should we define a threshold for determining whether a criterion is `inextricably linked'?

In challenging cases of alleged inherent discrimination, courts frequently revert to stating that allegations "can often be answered by asking the 'but for' question: but for the [claimant’s protected characteristic], would she have been treated more favourably?" \citep{Mruke}.\footnote{Note that the `but for' test is just one potential way of understanding the causation standard applied by courts in both the EU and the UK, and reflects the vast majority of outcomes in direct discrimination cases. However, neither UK nor EU courts have held that the concept of 'on grounds of' or 'because of' can be answered solely by reference to a `but for' test, including because the protected characteristic may form part of the background but not act as a causative factor (a point discussed further below). Our reasoning here thus contains some inevitable simplification. As noted above, nuance in the case law is one reason that a definitive mathematical 'test' for algorithmic direct discrimination cannot be laid down.}
This makes sense, given that the fundamental test is whether the outcome was \textit{on grounds of} the protected characteristic. Take each of the cases mentioned above. A claimant in a country with a sex-based retirement age could reasonably argue that he would have reached the retirement age \textit{but for} his sex;  a same-sex couple who live in a country which prohibits same-sex marriage could readily say that they would be married \textit{but for} their sexual orientation; a woman could reasonably argue that she would not be pregnant \textit{but for} her sex; and a Muslim woman could argue that she would not wear a headscarf \textit{but for} her religion. It therefore appears that while a probabilistic relationship between a criterion and a characteristic must be established in order for indirect discrimination to be made out, for \textit{direct} discrimination a \textit{deterministic relationship} is required.

The above examples are simple: each refers to a single criterion with an intuitive relationship with the protected characteristic. Such simplicity is likely to be more elusive in the algorithmic context. In particular, previous work has established that \textit{multiple} variables may constitute an inextricably linked proxy \textit{when used together} \citep{binns2023legal}. We will refer to this as a \textit{complex proxy}. To take a very basic example, a school might have previously operated on a boys-only basis but switched to a co-educational model in a particular year. Attendance at the school might not be inextricably linked to sex on its own, because after some time there may be plenty of female graduates. However, a requirement that a person both attended the school \textit{and} is above a certain age will be inherently discriminatory if the only people who can satisfy that requirement are those who attended the school when it was single-sex (i.e. male graduates). In the machine learning context, a range of criteria might be identified that, in combination, are inherently discriminatory. It remains to be seen how vast a range of complex proxies the courts will be willing to consider when identifying an inextricable link. 

The above approach – which essentially involves asking whether a criterion would apply to a person \textit{but for} their protected characteristic – also raises challenging philosophical questions. While counterfactual approaches towards measuring discrimination have been common in law and the social sciences \citep{kohler2018eddie} and, more recently, have been proposed in computer science literature \citep{pearl2016causal,kusner2017counterfactual,chiappa2019path, nabi2018fair}, analysing the causal effects of social categories in this way has been contested in the causal inference literature as well as the philosophy of social science. Some arguments against counterfactual accounts of discrimination are practical: if protected categories cannot be manipulated, we cannot empirically verify whether a protected characteristic was a cause of unfavourable treatment \citep{kohler2018eddie}. Other critiques of counterfactual reasoning are of a more theoretical nature, arguing that reducing protected characteristics to simplistic categories that can be manipulated in isolation (e.g., in causal graphs such as Figure~\ref{fig:causalexplanation}) fails to represent them meaningfully as social constructs \citep{kohler2018eddie,hu2020what,hanna2020towards,kasirzadeh2021counterfactuals}. In contrast, proponents of counterfactual analysis argue that it has the advantage of requiring auditors to make explicit the exact causes of unfavourable treatment, be it a direct effect of a relevant but simplistic measure of a protected characteristic or the consequences of a macro-level social structure \citep{bright2016causally,malinsky2018intervening,ross2023social}.

The law has some way to go to address such critiques. At present, it remains the case that courts will routinely deploy a hypothetical comparator to examine whether direct discrimination arose in a given case. To the extent that this paper proposes a means by which that case law can be understood from the perspective of algorithmic discrimination, we also adopt the approach taken by the courts while recognising potential methodological, philosophical, and political issues it gives rise to.

\subsection{Proxy Use}
Direct discrimination not only requires that the proxy be inextricably linked to a protected characteristic: it also has to be used, such that the individual is treated less favourably than they otherwise would have been. 

Courts often make reference to the ‘but for’ test here, too – although some caution is warranted. The fact that a protected characteristic "is a part of the circumstances in which the treatment complained of occurred, or of the sequence of events leading up to it, does not necessarily mean that it formed part of the ground, or reason, for that treatment" \citep{Ahmed}. To take a simple example, a female employee at a women-only gym who steals from a bag left in the changing rooms at her workplace and is consequently dismissed could perhaps say that she would not have been dismissed ‘but for’ having had access to the female changing rooms (an inherently discriminatory criterion) – but she would not have a good claim for direct discrimination, because her changing room access was not the reason for her dismissal. It was merely a background factor. A real-life example of this reasoning comes from the UK case of \textit{B v A}~\citep{BvA}. In that case, a female employee who had been in a romantic relationship with her manager was dismissed following the breakdown of the relationship. The employee made a claim of direct sex discrimination, arguing that she would not have been dismissed but for the fact that she was a woman. Her claim was unsuccessful. The Tribunal held that "[t]he dismissal occurred because of relationship breakdown, nothing more and nothing less than that"\citep{BvA}. In other words, the employee's sex was a background factor, but not a reason for the dismissal in itself. 

The best way to put it is that while the proxy criterion does not have to have been the only or even the main cause, it does have to have had a significant influence on the outcome \citep{Nagarajan}. While the courts have not explicitly designated this `significant influence' threshold as a bright-line test, scholars have identified that it best reflects the consistent approach in the jurisprudence \citep{binns2023legal}.

The concept of 'less favourable treatment' is a broad one. It includes, for example, refusal of entry to a restaurant or shop;  receipt of a smaller pension or lower pay; subjection to verbal abuse; or exclusion from certain educational opportunities \citep{FRA,Birmingham}. In the algorithmic context, the most likely form of less favourable treatment will be a poorer algorithmic score, and the implications that carries in the particular context. Other examples might include (for example) a large language model generating racist or sexist text \citep{RBS}.

Note that once it is shown that an inextricably linked proxy had a significant influence on a given claimant receiving a more disadvantageous outcome, there is (generally) no opportunity for the decision-maker to try to justify it by explaining why. This distinguishes the use of inextricably linked proxies from the use of proxies which are only weakly causally linked to protected characteristics, or linked by statistical association alone, and which thus fall within the realm of indirect discrimination law (unless caught by another form of direct discrimination). The use of the latter types of proxy may be justified by reference to reasonable and proportionate ends, whereas direct discrimination (as noted above) cannot be justified. For example, in \textit{Tests-Achats} \citep{Test-Achats}, the CJEU ruled that the use of gender in the determination of insurance premiums was incompatible with the principle of equal treatment, even though gender is genuinely useful in helping to predict insurance claims.

\section{Identifying Potential Inherent Direct Discrimination}
\label{sec:technicalanalysis}
The current leading approach for algorithmic fairness assessments is disaggregated analysis, in which a particular statistic (e.g., the proportion of predicted positives) is compared across groups \citep[e.g.,][]{weerts2023fairlearn}. While some types of disaggregated analyses have been argued to be consistent with the \textit{prima facie} evidence that is required for indirect discrimination \citep{wachter2021fairness}, such an effects-based analysis is insufficient for identifying and assessing inherent direct discrimination. In this section, we outline technical approaches and metrics that could be developed or applied to measure proxy capacity and proxy use in the context of inherent direct discrimination. 

\subsection{Identifying Potential Proxy Capacity}
From our legal analysis in the previous section, we can identify several conditions that shape our technical problem formulation. First, while it is unclear from case law how strong the relationship between the protected characteristic and its proxy must be, it is clear that the variables must at least be highly associated and possibly even deterministically related. Importantly, the inextricable link need not hold for \textit{all} protected groups. For example, in jurisdictions where same-sex couples cannot get married, marital status is inextricably linked to homosexuality but not heterosexuality: a straight person could be married, divorced, widowed, or never married, while a person who engages exclusively in same-sex relationships would never be married. 
Second, while it remains to be seen what the courts' stance is on complex proxies consisting of multiple variables, it is likely that at least a limited set of complex proxies would be considered.
Third, the link need not hold for \textit{all} members of a protected group  – it is sufficient if it holds for a subgroup (e.g., men aged 60 to 64, as in \textit{James v Eastleigh} or women who have attended a same-sex school). The technical problem formulation thus becomes: \textit{can we identify a (set of) variable(s) that are highly associated with (a subgroup of) a protected group}?

\subsubsection{Measuring Simple Proxy Capacity}
The most straightforward way of measuring potential proxy capacity is via measures of statistical association. Depending on the type of data (numerical, ordinal, binary, categorical), different measures could be appropriate (e.g., Pearson correlation, Spearman's rank correlation, mutual information). Measures of association are widely known and easy to compute. 

As explained above, the inextricable link need not hold for \textit{all} protected groups. Consequently, metrics that take into consideration \textit{all} categories of a protected characteristic may not be able to capture an inextricable link. For example, consider the variables \textit{sexual orientation} and \textit{marital status} in a jurisdiction where same-sex couples cannot get married. Computing association between these variables (e.g., via mutual information) could reveal an association, but the strength would be limited: while \textit{sexual orientation=`gay or lesbian'} would virtually exclusively co-occur with \textit{marital status=`never married'}, the co-occurrence of \textit{sexual orientation=`straight'} and the different categories of \textit{marital status} would not be deterministic. Considering categorical variables, we can learn more about the strength of a potential association by inspecting the contingency table of the two variables, which shows the frequencies of the co-occurrence of categories.

\paragraph{Example: proxy capacity in the \textit{Adult} Dataset}
We illustrate a proxy capacity analysis on the UCI \textit{Adult} dataset~\citep{adult}. This dataset was derived from the US Census of 1994 and contains demographic information of 48842 individuals. The \textit{Adult} dataset is commonly used in (fairness-aware) machine learning benchmarks, where the associated prediction task is to predict whether an individual earns more or less than \$50.000 per year.\footnote{We emphasise that, disconnected from a real-world use case, it is impossible to make conclusive claims regarding fairness or discrimination. For example, the protected status of characteristics differs across sectors, with the widest protection in employment. Similarly, measured proxy capacity can differ depending on which population the dataset was sampled from. A proxy for gender or race in one particular context may not be a proxy for a protected characteristic at a population level (or vice versa). For example, an insurer may have a particular type of customer that self-selects based on various factors, such as exposure to marketing and the policies offered. In this paper, we merely use the \textit{Adult} dataset for illustrative purposes.}

The \textit{Adult} dataset contains several variables that, depending on the sector in which a model trained on this dataset is applied, could fall within the material scope of EU law: \textit{sex}, \textit{race}, and \textit{age}. As discussed in Section~\ref{sec:legalcapacity}, protected characteristics such as race and gender are best viewed as multi-dimensional social constructs. Choosing one measurement over another can be more or less appropriate, depending on the context of an algorithmic decision-making system. In the case of the \textit{Adult} dataset, responses for \textit{race} are based on self-identification and \textit{sex} is worded to capture a person's biological sex (as opposed to gender). In the absence of a specific application, it is unclear whether these are appropriate measurements.

In this example, we test whether there are potential simple proxies for the \textit{sex} and \textit{race} variables in the dataset. As a measure of association, we use mutual information: an information-theoretic measure of dependence between variables that measures the extent to which knowing one variable reduces the uncertainty regarding the value of another variable. Mutual information varies between 0 (the variables are independent) and 1 (the variables are interchangeable).

\begin{table*}[ht]
    \centering
    \footnotesize
    \caption{Mutual information between two potential protected characteristics (\textit{sex} and \textit{race}) and other categorical features in the \textit{Adult} dataset.}
    \label{tab:mutualinfo}
    \begin{tabular}{l|r|r|r|r|r|r}
                & workclass & education	& marital-status & occupation &	relationship & native-country \\ \hline
         sex    & 0.013 & 0.005 & 0.112 & 0.099 & 0.271 & 0.002 \\
         race   & 0.007 & 0.001 & 0.012 & 0.012 & 0.017 & 0.094
    \end{tabular}
\end{table*}

None of the mutual information scores are close to 1, but the most likely proxy candidates are: \textit{relationship} as a proxy for \textit{sex} (0.271) and \textit{native-country} as a proxy for \textit{race} (0.094) (Table\ref{tab:mutualinfo}). We can further explore these potential proxies using contingency tables. 

The \textit{relationship} variable denotes the relationship the individual has to the householder. Considering \textit{sex} and \textit{relationship}, we observe a very clear pattern: virtually all instances for which \textit{relationship=Husband} we have \textit{sex=Male}, and vice versa, virtually all instances for which \textit{relationship=Wife} we have \textit{sex=Female} (Table~\ref{tab:contrelsex}). This analysis reveals that the \textit{relationship} variable has a strong proxy capacity for \textit{sex}. If, as we have suggested in Section~\ref{sec:legalcapacity}, the Court uses a counterfactual notion of `inextricable', a statistical proxy capacity measure must be accompanied by a plausible causal explanation. In the case of \textit{relationship} and \textit{sex}, there is a clear counterfactual connection: \textit{Husband} and \textit{Wife} are gendered definitions, implying that \textit{but for} their sex, a person would be classified differently. As such, these categories are almost certainly inextricably linked to sex.

\begin{table*}[ht]
    \centering
    \footnotesize
    \caption{A contingency table of the \textit{sex} and \textit{relationship} variables in the Adult dataset.}
    \label{tab:contrelsex}
    \begin{tabular}{l|r|r|r|r|r|r}
                &  Husband	& Not-in-family	& Other-relative & Own-child & Unmarried & Wife \\ \hline
         Female & 1 &	5870&	689	& 3376 &	3928 &	2328 \\ 
         Male &  19715	& 6713	& 817	& 4205	& 1197	& 3 \\ 
    \end{tabular}
\end{table*}

Considering the \textit{native-country} variable, we can identify a similar pattern: \textit{native-country=Laos} co-occurs exclusively with \textit{race=Asian-Pac-Islander} (Table~\ref{tab:contracenative}). However, in contrast to the \textit{relationship}, the court is less likely to accept this variable as an inextricable link. First, there is the problem of statistical significance: only 23 instances in the dataset have \textit{native-country=Laos}. Second, a counterfactual explanation is missing: would \textit{native-country} have been different {but for} \textit{race=Asian-Pac-Islander}?  In \textit{Jyske Finans}\citep{jyskefinans}, the CJEU has made it clear that it cannot be presumed that all citizens of a country are of a single ethnic origin.\footnote{"Ethnic origin cannot be determined on the basis of a single criterion but, on the contrary, is based on a whole number of factors, some objective and others subjective [...] As a consequence, a person’s country of birth cannot, in itself, justify a general presumption that that person is a member of a given ethnic group such as to establish the existence of a direct or inextricable link between those two concepts. Furthermore, it cannot be presumed that each sovereign State has one, and only one, ethnic origin." \citep[][paras 19-21]{jyskefinans}. Note that EU Council Directive 2000/43/EC of 29 June 2000 refers to "racial or ethnic origin" as a single concept~\citep{racialequalitydirective}.}
\begin{table*}[ht]
    \centering
    \footnotesize
    \caption{A contingency table of the \textit{race} and \textit{native-country} variables in the Adult dataset, including only the categories \textit{Laos} and \textit{United-States}.}
    \label{tab:contracenative}
    \begin{tabular}{l|r|r}
            &   Laos &	United-States \\ \hline 
    Amer-Indian-Eskimo	 & 0	& 452 \\
    Asian-Pac-Islander	 & 23	& 429 \\
    Black	             & 0	& 4269 \\
    Other                & 0    & 189 \\
    White                & 0    & 38493 \\
    \end{tabular}
\end{table*}

\subsubsection{Measuring Complex Proxy Capacity}
While association measures can be used to flag simple potential proxies, they are less likely to identify complex proxies. Many association measures rely on assumptions, particularly regarding the linearity of the relationship between the proxy and the protected characteristic. However, even a relatively simple model such as a decision tree can capture non-linear relationships. Moreover, measures of association are usually only suitable for measuring the relationship between two variables. Complex machine learning models could (unintentionally) rely on complex proxies: a set of features that together are predictive of a protected characteristic. In some cases, we can work around this limitation by creating a new variable that represents the intersection of multiple features. For example, consider the above example of a school that transitioned from being single-sex to co-educational after a particular year. We could transform two features, \textit{school\_attended} and \textit{years\_since\_graduation}, to devise a new variable \textit{attended\_singlesex\_school}, which will be highly associated with \textit{sex}. However, it can be difficult to anticipate the correct transformation in advance – especially if the associations identified by the machine learning model become more complex and less intuitive. One example would be geolocation data \citep{hu2019characterizing}. Various places in physical space may be inextricably linked with a protected characteristic; for instance, gender-segregated and disabled bathrooms, or places of religious worship. An algorithm for targeting adverts based on geolocation data, designed to optimise click-through rates, would naturally end up responding to any differences in responses based e.g. on gender, religion, or pregnancy status where these correspond with location. While Google and Apple's location-based advertising networks only offer coarse-grained geo-targeting, other advertising technologies, including Beacons, enable targeting to the nearest metre – more than sufficient to distinguish between male and female bathrooms in the same building \citep{senanayake2018accuracy}. Assessing proxy capacity for such forms of high-dimensional data may require substantial auxiliary data, and be highly context-dependent.

An alternative measure for capacity that circumvents these limitations, at least to some extent, is to consider how well a protected variable can be predicted from a proxy feature set. A similar approach was proposed by \citet{feldman2015certifying}, who set out a test for disparate impact that measures the extent to which a protected variable can be predicted from other features in a data set. For example, we can build a decision tree that predicts \textit{sex=Female} from \textit{school\_attended} and \textit{years\_since\_graduation}. We can then use any appropriate measure of predictive performance as a measure of proxy capacity. If we use the same model class (linear regression, decision tree, random forest, neural network etc.) for measuring capacity as the model that is deployed, the problem of restrictive assumptions of statistical association measures mentioned above can be mitigated. In other words, the model used to check for proxy capacity should be capable of identifying whatever kind of relationships - non-linear, interaction between variables - that may be present in the deployed model.

\subsubsection{Discovering Proxy Capacity}
In addition to an adequate measurement of proxy capacity, our technical problem formulation also points to the need for a search procedure that allows us to identify potential proxies that score high on capacity. Depending on the context, there could be multiple protected characteristics and various potential proxy features. Moreover, our legal analysis has shown that a proxy does not need to have the capacity for \textit{all} members of a protected group: it is sufficient if the proxy has capacity for a subgroup within a protected group. 

This is a typical use case for local pattern mining approaches, such as subgroup discovery~\citep{herrera2011overview} and exceptional model mining~\citep{duivesteijn2016exceptional}. These frameworks consist of several components, including the possible subgroup descriptors, the definition of a quality measure that defines the "interestingness" of a subgroup, and a search strategy. Considering potential proxy discovery, we could apply an exceptional model mining instance to search for subgroups in the dataset, defined by at least one protected characteristic, in which a potential proxy variable has a high capacity for a protected characteristic, possibly weighted by the coverage of the subgroup. As a complete search quickly becomes computationally expensive, heuristic search strategies such as beam search or evolutionary algorithms could be applied.

A challenge of computational analyses that attempt to discover potential proxy capacity is the problem of multiple comparisons. Considering multiple protected characteristics, multiple potential simple proxies, potential complex proxies, and subgroups within protected groups, the number of evaluations increases exponentially. As a result, there exists a substantial risk of a type I error: identifying potential proxy capacity in the data sample, due to chance, while the proxy does not have capacity in the population the sample was drawn from.

\subsection{Measuring Proxy Use}
The fact that a set of features could be a proxy for another feature does not directly imply that (1) the model uses them, and (2) the use results in the protected group being treated less favourably. For example, if the proxy variable is not predictive of the target variable, it is unlikely that the proxy will be used to make predictions. Additionally, if the input data contains other more informative features, a model that is penalised for complexity (e.g. via L1-regularization) may resort to these features over the potential proxy feature. Considering proxy use, the question we need to answer is: \textit{when does a (set of) variable(s) influence the predictions of a subgroup defined by at least one protected characteristic in such a way that we consider the subgroup to be treated less favourably}?

Identifying a causal relationship between a proxy and predictions constitutes an interventional analysis, in which we test whether an intervention on the proxy affects the predictions. Considering simple proxies, an interventional analysis is relatively straightforward if one has access to an API of the machine learning model. For each instance that is suspected to be directly discriminated against, we can simply determine the effect of changing the proxy variable on the output of the predictive model. If the counterfactual results in a different score or even a different classification (e.g., getting hired), this could provide evidence for illegitimate proxy use. Considering continuous variables, such as age or income, the effect could be visualised using an individual conditional expectation (ICE) plot \citep{goldstein2015peeking}, in which the outcome of the model (e.g., predicted score) is plotted against the range of potential feature values. 

Determining whether the observed effect would be considered 'less favourable' remains highly contextual. Claimants are most likely to be successful if they can show that the use of the proxy resulted in a worse outcome for them. For example, the change in the predicted score may exceed the decision threshold such that the decision changes to a less favourable outcome, such as being denied a benefit (e.g., a job interview in resume selection) or being subjected to a burden (e.g., a more thorough manual inspection in fraud detection). 

Interventional analyses become less straightforward when we consider complex proxies. For example, skin colour is strongly related to the colour of pixels in a photo. Specific combinations of subranges of pixel colour and position are therefore likely to have proxy capacity for a person's skin colour. However, this does not mean that the model uses that particular combination of pixels to make predictions. It is difficult to precisely determine the influence of pixels as a proxy of skin colour on the outcome of the model. Commonly used explanation techniques for deep neural networks, such as saliency maps \citep{simonyan2013deep}, indicate which pixels, if altered slightly, would result in the largest change in predicted probability. A saliency map thus indicates the sensitivity of the prediction to the pixels in a person's face but does not tell you whether the sensitivity is directly related to skin colour or perhaps to other aspects captured by the pixels unrelated to race. Instead, an interventional analysis of pixels as a proxy for skin colour would require a set of instances with differing `skin colours' but otherwise identical features. In other words, to perform an interventional analysis, we must be able to specify which values of the complex proxy correspond to social categories. Such precise specifications are particularly difficult for unstructured data, though difficulty will vary across applications. While meaningful counterfactuals of subtle indicators of an applicant's gender, such as writing style, are difficult to obtain, an interventional analysis would likely be able to identify resumes that are downgraded because they contain the word `women's'. Apart from technical challenges, disentangling a complex proxy from other characteristics associated with protected group membership invites theoretical critiques similar to those of counterfactual tests of (non-proxy) discrimination. 

\section{Discussion}
\label{sec:discussion}
Our work opens up several directions for future research.

From a legal perspective, we limit ourselves to setting out the principles necessary to examine how the established proposition that algorithms can directly discriminate should be understood in practice. Our legal analysis adds two elements to the existing scholarship  \citep{binns2023legal}. First, we propose that the assessment of inherent discrimination requires an examination of  (i) the \textit{inextricably linked nature} and (ii) the \textit{application} \textit{of} a criterion (or criteria) which is (or are) inextricably linked to a protected characteristic. This approach is widely accepted in EU discrimination law - but has never been translated into a framework for technical analysis. 

Second, and more significantly, our legal analysis suggests that the courts require a \textit{deterministic relationship} between a proxy and a protected characteristic in order to find that there is an inextricable link between them. Although this requirement emerges from an examination of the cases, it has never been explicitly spelled out in either the case law or (to our knowledge) in the legal literature. Most scholarly work thus far instead (implicitly) assumes that the relationship is a statistical one, despite pointing out the unprincipled nature of that approach \citep{IDDFoundations}. The apparent existence of a deterministic relationship between protected characteristics and inextricably linked proxies is a topic for further discussion in legal scholarship.

On the technical side, we have only been able to touch upon some of the potential approaches to show proxy use and capacity. Future work is needed to further develop and test these measures and approaches. 

Finally, future work could focus on other types of direct discrimination, particularly subjective discrimination, in the algorithmic context.

\section{Conclusion}
\label{sec:conclusion}
The vast majority of the literature on algorithmic discrimination has centred on the US disparate impact doctrine, resulting in a set of (quantitative) assessment measures and approaches most suited to this type of discrimination. In this paper, we set out to broaden this narrow view by studying inherent direct discrimination. Our proposed framework approaches evidence of inherent discrimination as two distinct problems: the proxy capacity test and proxy use test. Our legal analysis sets out how the courts have approached each of these steps. We then set out technical approaches that could be applied to establish their existence in practice. Subject to the algorithmic and corporate transparency challenges which victims of discrimination can face, any empirical evidence gathered in reliance on the framework proposed herein could be used to put forward a case of inherent direct discrimination in the context of algorithmic decision-making.

\section*{Research Ethics and Social Impact}
\label{researchethics}

\subsection*{Ethical Considerations Statement}
This work does not describe experiments with users and/or deployed systems and does not rely on sensitive user data. While we recognise the ethical and statistical shortcomings of the Adult dataset \citep{ding2021retiring}, it is already widely circulated and the identities of individuals in the data are unknown.

\subsection*{Researcher Positionality Statement}
The research, disciplinary backgrounds, and personal views of the authors have influenced this work. Some authors of this work were, at the time of conducting the research, employed as researchers in higher education institutions; another as a lawyer in the UK. The authors' experiences in legal and data science practice have shaped our work, specifically through the choice of the development of a framework with a practical focus rather than a merely academic contribution. As a result of this practical focus, we have sought to provide material that will help legal practitioners challenge algorithms in the court, and help machine learning practitioners avoid legal risk, with the overall aim of reducing the deployment of harmful algorithms. We appreciate that legal approaches can at best play but one part in a wider movement necessary to overcome discrimination through algorithms and wider society.

The authors occupy professional and socio-economic positions which substantially insulate us from the worst effects of discriminatory algorithms; as such, we lack the unique first-hand insights into algorithmic harms which come from those with lived experience. However, our work is motivated and informed through engagement with their testimony.

\subsection*{Adverse Impact Statement}
\label{researchethics:adverseimpact}
There are some ways in which our work, once published, could have adverse impact. First, our framework is intended to be of use to prospective claimants and algorithmic developers, seeking to identify preliminary ‘red flags’ in their models. However, our proposed tests could be misinterpreted, misused, or misconstrued by practitioners as a certification of nondiscrimination. Second, the need for litigators to make claims about protected characteristics to prove inherent discrimination might inadvertently reify them in problematic ways. For instance, by biologising constructs which are better understood as social and thus perpetuating, among others, harmful gender essentialism and racialisation.

\begin{acks}
Adams-Prassl gratefully acknowledges funding from the European Research Council under the European Union’s Horizon 2020 research and innovation programme (grant agreement No 947806).
\end{acks}

\bibliographystyle{ACM-Reference-Format}
\bibliography{references}

\appendix

\end{document}